%% file: main.tex
\documentclass[]{ceurart}

\usepackage{amsmath}
\usepackage{amssymb}
\usepackage{booktabs}
\usepackage{multirow}
\usepackage{graphicx}
\usepackage{array}
\usepackage{placeins}
\usepackage{xcolor}

\newif\ifanonymous
\anonymoustrue
\anonymousfalse
\usepackage{algorithm}
\usepackage{algpseudocode}

\newcommand{\base}{b}
\newcommand{\cand}{c}

\begin{document}

\copyrightyear{2026}
\copyrightclause{Copyright for this paper by its authors.
  Use permitted under Creative Commons License Attribution 4.0
  International (CC BY 4.0).}

\conference{GenAIECommerce'26: The Third Workshop on Agentic and Generative AI for E-Commerce, co-located with RecSys, September 28, 2026, Minneapolis, MN, USA}

\title{Distill Globally, Adapt Locally: Reasoning Distillation and Product-Type Test-Time Training for Scalable Trade-Up Recommendation}

\ifanonymous
  \author[1]{Anonymous Author(s)}
  \address[1]{Anonymous Organization}
\else
  \author[1]{Siliang Liu}[
    orcid=0009-0007-4561-7548,
    email=celineli@amazon.com,
  ]
  \cormark[1]
  \author[1]{Mohammad Ghasemi}[
    email=mohamgd@amazon.com,
  ]
  \author[1]{Sapan Patel}[
    email=sapanp@amazon.com,
  ]
  \author[1]{Amin Banitalebi-Dehkordi}[
    email=aminbt@amazon.com,
  ]
  \address[1]{Amazon Everyday Essentials Technologies}
  \cortext[1]{Corresponding author.}
\fi

\input{Sections/00_abstract}


\maketitle

\input{Sections/01_introduction}
\input{Sections/02_related_work}

\input{Sections/03_data}        
\input{Sections/04_system_overview}
\input{Sections/05_experiments}

\input{Sections/06_analysis}    
\input{Sections/08_conclusion}

\bibliography{references}

\appendix
\input{Sections/09_appendix}

\end{document}

%% file: Sections/00_abstract.tex
\begin{abstract}
Trade-up recommendation aims to identify higher-quality alternatives that preserve a customer's purchase intent while offering upgraded benefits through improved formulation, certifications, or brand positioning.
Although large language models (LLMs) can reason about these subtle distinctions, applying them directly to hundreds of millions of product pairs in e-commerce stores is operationally impractical.
We introduce a \emph{two-level} framework that distills LLM-derived reasoning into an efficient non-generative student and subsequently adapts its decision boundary to product-type-specific trade-up criteria.
At \textbf{Level~1}, a retrieval-augmented few-shot LLM teacher generates both structured relation labels and natural-language rationales.
These rationales are encoded and transferred to a compact, non-generative embedding-pair classifier through alignment and contrastive objectives.
At inference, the student consumes only two precomputed 768-dimensional product embeddings, requiring neither LLM calls nor text generation.
On a fixed human-annotated benchmark of $8{,}352$ pairs, a 15.5M-parameter four-class reasoning-distilled student achieves an AUC of $0.924$ (95\% CI $[0.918,0.929]$), improving over the four-class label-only student with the same shallow architecture (AUC $0.912$); rationale supervision provides little benefit when the same task is collapsed to binary labels.
At \textbf{Level~2}, we introduce product-type test-time training (PT-TTT), which uses few-shot demonstrations as gradient-based supervision to optimize lightweight category-specific adapters over the frozen student model.
PT-TTT improves AUC from $0.924$ to $0.941$ and average precision from $0.920$ to $0.940$ without serving-time LLM inference.
On a 100K-pair proxy catalog, inference with the distilled student on a single eight-GPU machine is approximately $5{,}000\times$ faster and has an estimated cost approximately $10{,}000\times$ lower than direct LLM inference on the same workload.
\end{abstract}

%% file: Sections/01_introduction.tex
\section{Introduction}

\begin{figure}[t]
\centering
\includegraphics[width=\linewidth]{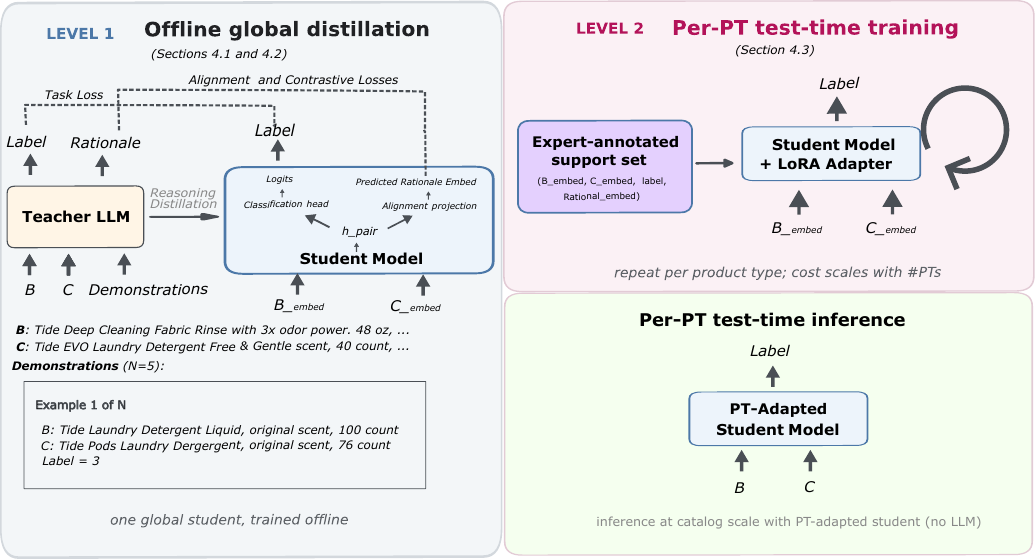} 
\caption{Two-level framework. \textbf{Level~1}: a frontier LLM teacher supervises a discriminative global student through reasoning-guided distillation.
\textbf{Level~2}: the \emph{same} per-product-type support sets that served as prompt context for the teacher are reused as test-time \emph{gradient} supervision for a lightweight per-category adapter, applied once per product type before catalog-scale scoring.
\textbf{Inference}: catalog-scale scoring with the PT-adapted student, with no LLM calls.}
\label{fig:ttt-concept}
\end{figure}

Trade-up recommendation identifies higher-quality alternatives that preserve a customer's shopping intent while offering additional benefits in attributes such as formulation, certifications, materials, or brand positioning.
At catalog scale, this requires distinguishing meaningful upgrades from same-tier substitutes, variants, pack-size changes, and incompatible products.

Given a base product $b$ and candidate $c$, we predict whether $c$ is a valid trade-up for $b$.
Unlike substitutability, variant, and complementarity relations ~\cite{mcauley2015inferring,herrerovidal2024variant,yamasaki2025complementary}, trade-up is directional: the candidate must preserve the shopping mission while providing evidence of an added benefit beyond price, quantity, or other superficial variation.

Large language models (LLMs) provide useful semantic supervision for recommendation and product relations~\cite{geng2022p5,bao2023tallrec,xi2024kar,wu2024llmrecsurvey}, and retrieval augmentation can ground their decisions with task-specific examples~\cite{lewis2020rag}.
However, applying LLM inference at e-commerce catalog scale is operationally impractical.
For a large e-commerce store, scoring the catalog can involve hundreds of millions of pairs, costing millions of dollars and delaying recommendation availability by weeks or months.

We propose a robust and scalable learning-based solution capable of identifying trade-up products.
To achieve this, we integrate a two-level reasoning framework to distill nuanced trade-up relations across product types. 
Our contributions are as follows:
\begin{itemize}
    \item We formulate trade-up identification as a directional, product-type-dependent relation classification problem designed for catalog-scale scoring without serving-time LLM inference.
    \item We introduce a reasoning-distillation framework that transfers fine-grained relation labels and rationale semantics from a retrieval-augmented LLM teacher to a compact, non-generative student (Sections~\ref{ssec:arch} and~\ref{ssec:training}).
    \item We introduce Product-Type Test-Time Training (PT-TTT), which uses product-type-specific demonstrations to fit lightweight adapters over the frozen student, improving predictive performance while preserving efficient large-scale inference (Section~\ref{ssec:ttt-method}).
\end{itemize}

%% file: Sections/02_related_work.tex
\section{Related Work}
\label{sec:related_work}

\paragraph{Product relations and LLM supervision.}
Item-to-item relation modeling has primarily addressed substitutability,
complementarity, compatibility, and product variants using behavioral,
textual, and graph signals~\cite{mcauley2015inferring}; a related line uses
fine-grained, ingredient-based product attributes for explainable e-commerce
recommendation~\cite{liu2024beauty}.
Formally, we model trade-up as a directional relation from a base product \(b\) to a candidate \(c\), where \(c\) must preserve the underlying shopping mission while providing an additional customer-relevant benefit.
We therefore treat trade-up identification as catalog-level relation classification upstream
of candidate ranking, rather than as an end-to-end personalized recommendation
problem.

LLMs have been used as recommenders, semantic feature generators, rerankers,
and annotators~\cite{geng2022p5}. Retrieval-augmented prompting can further
condition their predictions on task-specific demonstrations
\cite{lewis2020rag}. In our setting, the LLM serves only as an offline semantic
supervisor to produce a fine-grained product-relation label and a textual
rationale for each pair. This design targets a different operating regime from
LLM-based online recommendation, as the deployed scorer must process
hundreds of millions of pairs without generative inference. Moreover, the
teacher is accessed as a black box, precluding conventional distillation from
teacher logits or hidden states.

\paragraph{Rationale-guided representation distillation.}
Classical knowledge distillation transfers output distributions from a teacher
to a compact student~\cite{hinton2015distilling}; subsequent work matches
intermediate features~\cite{romero2015fitnets} or preserves relational geometry
through contrastive objectives~\cite{tian2020crd}. Reasoning-distillation
methods extend the supervision signal with explanations or chain-of-thought
traces. Distilling step-by-step jointly trains students on labels and
rationales~\cite{hsieh2023distilling}, while SCOTT introduces consistency
objectives between generated reasoning and predictions~\cite{wang2023scott}.
These methods predominantly target generative students that emit or condition
on rationale tokens.

Our student is an embedding-only pair classifier and does not decode text.
We instead encode the teacher rationale into a fixed-dimensional
representation and transfer it to the student pair embedding through
pointwise alignment and contrastive relational matching. The projection and
auxiliary losses are used only during training. This places our method between
rationale distillation and representation distillation. The teacher provides
natural-language supervision, but the transferred object is the geometry of
the discriminative pair representation rather than a generated reasoning
sequence.
We use the term \emph{rationale-guided representation distillation} because
generated rationales need not faithfully expose the teacher's internal
decision process~\cite{turpin2023unfaithful}. Accordingly, we treat them as
auxiliary semantic targets and evaluate their contribution separately from
model capacity and label granularity. This distinction is consequential in our
results: rationale supervision improves human-label alignment only when the
student retains the teacher's fine-grained relation structure, rather than
collapsing all non-trade-up cases into a single binary class.

\paragraph{Test-time product type-based adaptation.}
Test-time training adapts a trained model on the target distribution before
prediction~\cite{sun2020ttt}; related test-time adaptation methods update only a 
restricted parameter subset under distribution shift
\cite{wang2021tent}. Our setting is additionally connected to supervised
few-shot adaptation and meta-learning~\cite{finn2017maml}, since each product
type has a small labeled demonstration set. Recent work shows that such
demonstrations can be converted from in-context examples into gradient
supervision at inference time~\cite{akyurek2025ttt}. Parameter-efficient
finetuning methods, including LoRA~\cite{hu2022lora}, make this adaptation feasible
without modifying the shared backbone.

PT-TTT reuses the teacher's product-type-specific demonstrations as a support
set for adapting a low-rank module over the frozen student. Because the support
examples are labeled, the method is more precisely supervised few-shot
test-time adaptation than conventional unsupervised test-time training. In our system,
one adapter is optimized per product type
and amortized over all pairs in that category, rather than optimized per query.
This preserves the computational advantage of the distilled scorer while
allowing category-specific trade-up criteria to modify the local decision
boundary.

Overall, prior work studies semantic product relations, rationale
distillation, and test-time adaptation largely as separate problems. Our
framework combines them through a division of labor: rationale-guided
distillation shapes a globally shared pair representation, while PT-TTT
specializes its decision boundary using category-local supervision. The
contribution is this coupling under catalog-scale constraints, without
serving-time LLM inference or per-pair adaptation.

%% file: Sections/03_data.tex
\section{Task Setup and Data Preparation}
\label{sec:data}

\subsection{Trade-up Definition}
Given a base product $\base$ and candidate $\cand$, we predict a score $s(\base,\cand)\in[0,1]$ indicating whether $\cand$ is a valid trade-up of $\base$.
A valid trade-up must preserve the same shopping intent while providing an additional benefit, such as stronger brand positioning, improved formulation, relevant certifications, or better materials.
Differences in quantity, flavor, or packaging alone do not constitute a trade-up.

\subsection{Label Sources and Corpora}
\paragraph{Expert-annotated data and golden benchmark.} 

We construct an expert-annotated corpus of $17{,}200$ product pairs across 29 product types.
Within each type, selected products are exhaustively paired and assigned a fixed $\base$--$\cand$ ordering under a four-class taxonomy:
(1) similar/same tier;
(2) $\base$ is a trade-up of $\cand$;
(3) $\cand$ is a trade-up of $\base$; or
(4) incompatible/not meaningfully comparable.
We use $8{,}848$ pairs as the teacher demonstration and PT-TTT support pool and reserve the remaining $8{,}352$ pair-disjoint examples as the held-out golden benchmark.
For binary evaluation, class~3 is positive and classes~1, 2, and~4 are negative.
Golden labels are used only for final evaluation and analysis.

\paragraph{Silver supervision.}
Starting from $12{,}830$ products in the Amazon--Walmart dataset ~\cite{hpi_amazon_walmart}, we construct within-product-type candidate pairs and retain semantically related pairs based on product-title and description embeddings, filtering out pairs unlikely to represent the same shopping intent.
This yields $1{,}019{,}241$ candidate pairs across the same 29 product types.

For each retained pair, a frontier LLM teacher, accessed as a black box, receives the product descriptions, product-type-specific trade-up criteria, and five dynamically retrieved expert demonstrations, and generates a four-class relation label and a concise rationale.
These outputs constitute the silver supervision for global student training.
The corpus is split 90/10 into training and validation sets, stratified by product type and relation class.
Additional details on demonstration retrieval, teacher prompting, and rationale encoding are provided in
Appendix~\ref{app:teacher-details}.

%% file: Sections/04_system_overview.tex
\section{System Overview}

Figure~\ref{fig:ttt-concept} shows an overview of the two-level pipeline: offline reasoning distillation into a frozen embedding-only student in Level~1 (Section~\ref{ssec:arch} and \ref{ssec:training}), then per-product-type test-time training before scoring in Level~2 (Section~\ref{ssec:ttt-method}).

\subsection{Embedding-Pair Student Architecture}
\label{ssec:arch}

The student is a lightweight product-pair classifier that consumes precomputed base and candidate embeddings
$e_b,e_c\in\mathbb{R}^{d}$, with $d=768$.
It uses separate base and candidate branches together with an ordered base-to-candidate branch, followed by an MLP classification head.
No product text, rationale, or LLM output is required at inference.

The base and candidate embeddings are processed by separate parameterized branches. 
Because each product is represented by a single embedding, the self-attention operations act on length-one sequences and therefore have trivial attention weights.
They therefore act as learned role-specific transformations
rather than performing content-dependent attention over tokens.
A third, ordered base-to-candidate branch provides an additional asymmetric transformation. 
Because each side contains a single embedding, its attention
softmax is identically one and does not perform content-dependent selection.
Directionality instead arises from the ordered base/candidate roles, separate branch parameters, and the ordered pair representation.

Let $h_b^{(L_s)}$ and $h_c^{(L_s)}$ denote the final outputs of the
base and candidate branches, respectively, and let $z^{(L_x)}$ denote
the final output of the ordered base-to-candidate branch.
The three branch outputs are concatenated as
\begin{equation}
h_{\mathrm{pair}}
=
\bigl[
h_b^{(L_s)};
h_c^{(L_s)};
z^{(L_x)}
\bigr]
\in\mathbb{R}^{3d},
\label{eq:hpair}
\end{equation}
and an MLP maps $h_{\mathrm{pair}}$ to binary or four-class logits.

\paragraph{Reasoning projection.}
During distillation, a training-only projection maps the pair representation to the rationale-embedding space:
\begin{equation}
\hat r_i
=
W_{\mathrm{align}}h_{\mathrm{pair}}^{(i)}
+
b_{\mathrm{align}},
\qquad
\hat r_i\in\mathbb{R}^{d_r}.
\label{eq:proj}
\end{equation}
The projection is used only by the auxiliary distillation objectives and is discarded at inference.
Full branch equations, layer dimensions, parameter counts, and architecture hyperparameters are provided in Appendix~\ref{app:student-arch}.




\subsection{Student Model Training Objectives}
\label{ssec:training}

\paragraph{Task supervision.}
We consider binary and four-class supervision.
In binary mode, class~3 (candidate is a trade-up of the base) is treated as positive, while classes~1, 2, and~4 are collapsed into the negative class:
\[
y_i^{(2)}=\mathbb{I}[y_i^{(4)}=3].
\]
The binary model is optimized using weighted focal binary
cross-entropy~\cite{lin2017focal}.
In four-class mode, we preserve the teacher's original relation taxonomy and optimize weighted cross-entropy.
We denote either objective by $\mathcal{L}_{\mathrm{task}}$.

\paragraph{Rationale alignment.}
Because the black-box teacher exposes neither logits nor hidden states, we use its natural-language rationales as auxiliary semantic supervision.
Let $r^{(i)}$ be the encoded teacher rationale and $\hat r_i=W_{\mathrm{align}}h_{\mathrm{pair}}^{(i)}+b_{\mathrm{align}}$ the student projection defined in Eq.~\ref{eq:proj}.
We minimize
\begin{equation}
\mathcal{L}_{\mathrm{align}}
=
\frac{1}{N}
\sum_{i=1}^{N}
\left\|
\hat r_i-r^{(i)}
\right\|_2^2 .
\label{eq:align}
\end{equation}




\paragraph{Contrastive distillation.}
Pointwise alignment does not explicitly preserve relationships among different pair--rationale examples.
We therefore complement it with a contrastive objective $\mathcal{L}_{\mathrm{con}}$ combining InfoNCE~\cite{oord2018cpc,tian2020crd}, which contrasts each student projection against in-batch rationale embeddings, and a relational KL term ~\cite{park2019relational} that transfers pairwise similarity structure.









\paragraph{Full objective.}
The global student is trained with

\begin{equation}
\mathcal{L}
=
\mathcal{L}_{\mathrm{task}}
+
\lambda_{\mathrm{align}}\mathcal{L}_{\mathrm{align}}
+
\lambda_{\mathrm{con}}\mathcal{L}_{\mathrm{con}}.
\label{eq:total}
\end{equation}

The rationale projection and auxiliary distillation objectives are used only during training and discarded at inference.
For the four-class model, the trade-up score is the softmax probability of class~3 (candidate is a trade-up of the base):
$s(\base,\cand)=\operatorname{softmax}(g)_3$.
Full loss definitions and optimization hyperparameters are provided in
Appendix~\ref{app:loss-details}.





\subsection{Product-Type Test-Time Training (PT-TTT)}
\label{ssec:ttt-method}

The globally distilled student uses a shared scoring function across product types, whereas trade-up criteria can differ substantially by category: evidence that makes a battery a trade-up differs from that of a moisturizer (formulation, certifications) or pet food (ingredient
sourcing). 
We therefore add a second level of specialization inspired
by test-time training~\cite{sun2020ttt,akyurek2025ttt}. 
Because these support examples carry expert labels, the procedure is more precisely a supervised few-shot test-time adaptation method than conventional unsupervised test-time training.

\paragraph{Product-type adaptation.}
Let $\theta_0$ denote the frozen global student and $S_t=\{(e_b^{(i)},e_c^{(i)},y^{(i)},r^{(i)})\}_{i=1}^{K}$ the support set for product type $t$.
We fit lightweight LoRA adapters~\cite{hu2022lora} in the classification head and reasoning-projection layer while keeping the original student parameters frozen.

The adapted model is denoted
$\theta_t=\theta_0\oplus\phi_t$.
The reasoning-guided adaptation objective is
\begin{equation}
\mathcal{L}_{\mathrm{TTT}}(S_t)
=
\mathcal{L}_{\mathrm{task}}(S_t)
+
\lambda_{\mathrm{reason}}
\frac{1}{K}
\sum_{i=1}^{K}
\left\|
\hat r_i-r^{(i)}
\right\|_2^2 .
\label{eq:ttt}
\end{equation}

The label-only variant sets $\lambda_{\mathrm{reason}}=0$, whereas the reasoning-guided variant uses $\lambda_{\mathrm{reason}}=0.1$ and reuses the rationale-alignment signal from Eq.~\ref{eq:align}.
We omit the contrastive objective during adaptation because the small, product-type-homogeneous support sets provide few reliable in-batch negatives.
Algorithm~\ref{algo:pt_ttt} summarizes the complete adaptation-and-scoring
procedure.

\begin{algorithm}
   \small
   \caption{\small PT-TTT: test-time adaptation and scoring for product type $t$}
   \label{algo:pt_ttt}
   \begin{algorithmic}[1]
      \State \texttt{\textcolor{blue}{\# Frozen global student $\theta_0$; PT-specific adapter parameters $\phi$}}
      \State \texttt{model = load\_global\_student()} \Comment{$\theta_0$ remains frozen}
      \State \texttt{inject\_lora(model, rank=8, alpha=16)}
      \State \texttt{$\phi$\_init = save\_adapter\_initialization(model)}
      \State

      \Function{adapt\_and\_score}{\texttt{support\_t, query\_t, use\_reason}}
         \State \texttt{\textcolor{blue}{\# Every PT starts from the same adapter initialization}}
         \State \texttt{restore\_adapter(model, $\phi$\_init)}
         \State \texttt{opt = AdamW(adapter\_params(model), lr=1e-3)}
         \State
         \State \texttt{model.train()}
         \For{\texttt{step in 1..50}}
            \State \texttt{logits, h\_pair = model(support\_t.base, support\_t.cand)}
            \State \texttt{L = task\_loss(logits, support\_t.label)}
            \If{\texttt{use\_reason}}
               \State \texttt{r\_hat = alignment\_proj(h\_pair)}
               \State \texttt{L = L + 0.1 * mse(r\_hat, support\_t.reason\_emb)}
            \EndIf
            \State \texttt{opt.zero\_grad()}
            \State \texttt{L.backward()}
            \State \texttt{opt.step()}
         \EndFor
         \State
         \State \texttt{\textcolor{blue}{\# Score unlabeled query pairs with the PT-adapted model}}
         \State \texttt{model.eval()}
         \State \texttt{with no\_grad():}
         \State \hspace{\algorithmicindent}\texttt{logits = model(query$_t$.base, query$_t$.cand)}
         \State \hspace{\algorithmicindent}\texttt{p = softmax(logits)[:, tradeup\_class]}
         \State \Return \texttt{p}
         \Comment{Discard $\phi_t$; the next PT restarts from $\phi_{\mathrm{init}}$}
      \EndFunction
   \end{algorithmic}
\end{algorithm}

Each product type starts from the same global model and adapter initialization; adaptation is performed once per product type and amortized over all query pairs in that category.
Additional LoRA parameterization and implementation details are provided in Appendix~\ref{app:ttt-optimization}.

%% file: Sections/05_experiments.tex
\section{Experiments}

\subsection{Setup}
\label{ssec:data}

We conduct all experiments using the corpora described in Section~\ref{sec:data}. Specifically, the models are trained on $1{,}019{,}241$ teacher-annotated (\emph{silver}) pairs, partitioned using a stratified 90/10 training and validation split, and evaluated on a fixed human-annotated (\emph{golden}) benchmark comprising $8{,}352$ pairs. 
We compare shallow and deep student capacities; the shallow reasoning model
contains $15.5$M parameters and the corresponding deep model $65.9$M.
Exact layer configurations are provided in Appendix~\ref{app:student-arch}.

\paragraph{Evaluation.}
Models and checkpoints are selected using the silver validation split; the $8{,}352$-pair golden benchmark is held out for final evaluation.
We report AUC and average precision (AP) as the primary threshold-independent metrics~\cite{anelli2021elliot}, together with F1, precision, and recall.
Unless otherwise noted, thresholded metrics use operating points selected on validation data and applied unchanged to the golden benchmark.
Golden-set confidence intervals use $2{,}000$ bootstrap resamples, with paired bootstrap resampling for model comparisons.
Full evaluation and statistical details are provided in Appendix~\ref{app:evaluation-details}.


\subsection{Main Results}
\label{ssec:reason-result}

Table~\ref{tab:reason-ablation} reports performance on the fixed $8{,}352$-pair golden benchmark for the LLM teacher, label-only students, and reasoning-distilled students. 
Figure~\ref{fig:headline} summarizes the principal comparison together with bootstrap confidence intervals. 
Additional results for global reasoning distillation are provided in Appendix~\ref{app:level1}.
Taken together, these results support three main conclusions.


\begin{table}[t]
\centering
\caption{\textbf{Golden benchmark results} ($n{=}8{,}352$), with 95\% bootstrap CIs on AUC.
F1/P/R are evaluated on the golden benchmark using the best-F1 operating threshold selected on the validation set for each model.
The shallow-model block provides a same-architecture comparison across label granularity (binary vs.\ four-class) and rationale supervision (label-only vs.\ +Reason).
The best observed result is the \emph{shallow} $15.5$M reasoning-distilled four-class student.
Row groups: teacher; shallow-architecture comparisons; larger/deep students.
}
\label{tab:reason-ablation}
\small
\begin{tabular}{lcccccc}
\toprule
Model & Params & AUC (95\% CI) $\uparrow$ & AP $\uparrow$ & F1 $\uparrow$ & Prec. $\uparrow$ & Rec. $\uparrow$ \\
\midrule

LLM teacher \textit{no demo}
& ---
& ---
& ---
& 0.724
& 0.918
& 0.597 \\

LLM teacher \textit{RAG, 5 demo}
& ---
& ---
& ---
& 0.749
& 0.967
& 0.610 \\

\midrule

Label-only (binary, shallow)
& 14.3M
& 0.911,{\scriptsize[.905,.917]}
& 0.916
& 0.834
& 0.812
& 0.857 \\

+ Reason (binary, shallow)
& 15.5M
& 0.911,{\scriptsize[.905,.918]}
& 0.914
& 0.836
& 0.822
& 0.850 \\

Label-only (4-class, shallow)
& 14.3M
& 0.912,{\scriptsize[.906,.918]}
& 0.916
& 0.836
& 0.820
& 0.853 \\

\textbf{+ Reason (4-class, shallow)}
& \textbf{15.5M}
& \textbf{0.924},{\scriptsize[.918,.929]}
& \textbf{0.920}
& \textbf{0.843}
& 0.829
& \textbf{0.858} \\

\midrule

Label-only (binary, deep)
& 64.7M
& 0.887,{\scriptsize[.881,.893]}
& 0.889
& 0.805
& 0.803
& 0.807 \\

+ Reason (binary, deep)
& 65.9M
& 0.907,{\scriptsize[.901,.913]}
& 0.907
& 0.833
& 0.857
& 0.810 \\

+ Reason (4-class, deep)
& 65.9M
& 0.911,{\scriptsize[.905,.917]}
& 0.902
& 0.832
& 0.806
& 0.860 \\

\bottomrule
\end{tabular}
\end{table}

\paragraph{(1) The best student is shallow and reasoning-distilled.}
The shallow four-class reasoning-distilled student achieves AUC $0.924$ ($95\%$ CI $[0.918,0.929]$), compared with $0.912$ for the four-class label-only model using the same shallow architecture.
It also exceeds the larger deep label-only model (AUC $0.887$).

\paragraph{(2) Rationale and label granularity interact.}
Within the same shallow architecture, adding rationale supervision to the binary model leaves AUC unchanged at the reported precision ($0.911$ vs.\ $0.911$), with only small changes in AP and thresholded metrics.
In contrast, combining rationale supervision with four-class labels improves AUC from $0.912$ to $0.924$; using the unrounded predictions, the paired difference is $\Delta_{\mathrm{AUC}}=+0.013$ (95\% CI $[+0.010,+0.016]$).
Thus, the observed reasoning benefit depends on retaining the teacher's fine-grained relation structure.

\paragraph{(3) The student improves F1 relative to the evaluated teacher configuration.}
The retrieval-augmented teacher with five demonstrations achieves precision $0.967$, recall $0.610$, and F1 $0.749$.
The distilled student reaches precision $0.829$, recall $0.858$, and F1 $0.843$ on the same benchmark, primarily by recovering recall.
This comparison is specific to the evaluated retrieval-augmented teacher configuration and should not be interpreted as evidence that the student generally outperforms frontier LLMs.




\subsection{PT-TTT}
\label{ssec:ttt}



We use the best Level-1 model, the shallow 15.5M four-class reasoning-distilled student (AUC $0.924$), as the frozen global student.
We evaluate PT-TTT with support sizes $K\in\{4,8,16,32\}$ under label-only and reasoning-guided adaptation.
Implementation details are given in Section~\ref{ssec:ttt-method} and Appendix~\ref{app:ttt}.

\begin{table}[t]
\centering
\caption{\textbf{PT-TTT} on the golden benchmark ($n{=}8{,}352$). $K$ is the per-PT support size. 
The \emph{global} baseline ($K{=}0$) corresponds to the frozen shallow 15.5M-parameter reasoning-distilled four-class student, evaluated without test-time adaptation, the best-performing model from Table~\ref{tab:reason-ablation}, denoted \mbox{+ Reason (4-class)}.
\textbf{Label-only} and \textbf{Reasoning} are the two adaptation objectives, fitting the \emph{same} LoRA adapter without vs. with the rationale-alignment term ($\lambda_{\text{reason}}{=}0$ vs. $>0$).
Both improve threshold-independent AUC/AP monotonically over the global model; the lift is recall-driven, and the two objectives are near-identical (rationale adds little at test time). 
All rows report F1/precision/recall at a single fixed operating threshold ($0.45$) held constant across $K$, so the frozen global model and every adapted model are compared at the same operating point. }

\label{tab:ttt}
\small
\begin{tabular}{llccccc}
\toprule
Adapt. & $K$ & AUC $\uparrow$ & AP $\uparrow$ & F1 $\uparrow$ & Prec. $\uparrow$ & Rec. $\uparrow$ \\
\midrule
\emph{global} & 0 & 0.924 & 0.920 & 0.839 & 0.862 & 0.817 \\
\midrule
\multirow{4}{*}{Label-only}
 & 4  & 0.929 & 0.924 & 0.839 & 0.841 & 0.837 \\
 & 8  & 0.937 & 0.932 & 0.844 & 0.810 & 0.881 \\
 & 16 & 0.940 & 0.936 & 0.860 & 0.844 & 0.876 \\
 & 32 & 0.940 & 0.938 & 0.856 & 0.846 & 0.867 \\
\midrule
\multirow{4}{*}{Reasoning}
 & 4  & 0.925 & 0.920 & 0.829 & 0.838 & 0.821 \\
 & 8  & 0.937 & 0.931 & 0.847 & 0.818 & 0.878 \\
 & 16 & 0.940 & 0.936 & 0.859 & 0.844 & 0.874 \\
 & 32 & \textbf{0.941\,{\scriptsize(+.017)}} & \textbf{0.940\,{\scriptsize(+.020)}} & \textbf{0.856\,{\scriptsize(+.017)}} & 0.845\,{\scriptsize(--.017)} & \textbf{0.867\,{\scriptsize(+.050)}} \\
\bottomrule
\end{tabular}
\end{table}

\paragraph{Adapter Gains.} 
PT-TTT improves the globally distilled student across the evaluated support budgets (Table~\ref{tab:ttt}).
At $K{=}32$, reasoning-guided adaptation increases golden AUC from $0.924$ to $0.941$ ($+0.017$) and AP from $0.920$ to $0.940$ ($+0.020$).
Label-only PT-TTT reaches nearly the same AUC ($0.940$).
Performance largely plateaus between $K{=}16$ and $K{=}32$; the complete support-size curves are provided in Appendix~\ref{app:ttt-support}.



\paragraph{Matched human-supervision control.}
PT-TTT is directly optimized on expert-labeled support examples, whereas the global student is trained on LLM-generated silver supervision.
To separate the effect of direct expert optimization from that of product-type specialization, we train a category-agnostic pooled LoRA adapter using exactly the same expert-labeled support examples as PT-TTT.
At $K{=}32$, both approaches therefore use $928$ support examples across the 29 product types.
As shown in Table~\ref{tab:ttt-pooled-control}, pooled adaptation improves AUC from $0.924$ to $0.929$, whereas product-type-specific label-only adaptation reaches $0.940$.
Thus, direct optimization on expert labels explains part, but not all, of the PT-TTT improvement.

\begin{table}[t]
\centering
\caption{\textbf{Matched human-supervision control for PT-TTT.}
The pooled and per-product-type adapters use the same expert-labeled support examples ($K{=}32$ per product type; $928$ examples in total across 29 product types) and start from the same frozen global student.
The pooled baseline fits one category-agnostic LoRA adapter to the union of the support examples, whereas PT-TTT independently fits one adapter per product type.
F1/P/R use the same fixed threshold ($0.45$) as Table~\ref{tab:ttt}}
\label{tab:ttt-pooled-control}
\small
\begin{tabular}{lcccccc}
\toprule
Method &
Total expert support &
AUC $\uparrow$ &
$\Delta$ vs.\ global &
AP $\uparrow$ &
F1 $\uparrow$ &
Prec./Rec. $\uparrow$ \\
\midrule

Global student
& 0
& 0.924
& ---
& 0.920
& 0.839
& 0.862 / 0.817 \\

Pooled LoRA (label-only)
& 928
& 0.929
& +0.005
& 0.926
& 0.845
& 0.852 / 0.839 \\

PT-TTT (label-only)
& 928
& 0.940
& +0.016
& 0.938
& 0.856
& 0.846 / 0.867 \\

PT-TTT (+ Reason)
& 928
& \textbf{0.941}
& \textbf{+0.017}
& \textbf{0.940}
& 0.856
& 0.845 / 0.867 \\

\bottomrule
\end{tabular}
\end{table}

\paragraph{Within-product-type ranking control.}
Pooled AUC can improve if score distributions shift differently across product types, even when within-product-type discrimination remains unchanged.
We therefore compute AUC independently within each product type and report the unweighted macro average across the 29 types.
As shown in Table~\ref{tab:ttt-perpt}, macro PT-AUC increases from $0.910$ for the global student to $0.914$ for pooled LoRA and $0.925$ for PT-TTT.
The larger within-type gain from PT-TTT indicates improved product-type-level discrimination rather than only cross-category score rescaling.

\begin{table}[t]
\centering
\caption{\textbf{Pooled and within-product-type AUC.}
Pooled AUC is computed jointly over all golden benchmark pairs.
Macro PT-AUC is the unweighted average of AUC computed independently
within each product type. Improvement in Macro PT-AUC indicates improved
within-product-type discrimination rather than only cross-category
score rescaling.}
\label{tab:ttt-perpt}
\small
\begin{tabular}{lccc}
\toprule
Method &
Pooled AUC $\uparrow$ &
Macro PT-AUC $\uparrow$ &
$\Delta$ Macro PT-AUC \\
\midrule
Global student
& 0.924
& 0.910
& --- \\

Pooled LoRA (label-only)
& 0.929
& 0.914
& +0.004 \\

PT-TTT (label-only)
& 0.940
& 0.925
& +0.015 \\

\bottomrule
\end{tabular}
\end{table}

\paragraph{Rationale reuse during adaptation.}
Reasoning-guided and label-only PT-TTT perform similarly at moderate support sizes. At $K{=}8$ and $K{=}16$, their AUCs are identical at the reported precision, and at $K{=}32$ they reach $0.941$ and $0.940$, respectively.
Thus, the Level-2 improvement appears to arise primarily from
product-type-specific adaptation rather than from reusing rationale
supervision during adaptation.

\subsection{Scalability}
\label{ssec:scalability}
As a small-scale proxy for full-catalog inference, we benchmark the distilled student on approximately 100K randomly sampled product pairs from the Amazon--Walmart dataset~\cite{hpi_amazon_walmart}.
Even on this small-scale benchmark, running the distilled student on a single eight-GPU machine yields approximately a $5{,}000\times$ speedup and a $10{,}000\times$ reduction in estimated inference cost relative to direct LLM inference on the same workload.
This scalability is enabled by separating training-time reasoning supervision from serving-time prediction: the rationale encoder, projection head, and auxiliary distillation objectives are used only during training, whereas inference requires only the compact pair-classification model and precomputed product embeddings. 
Figure~\ref{fig:widget} in Appendix~\ref{app:deployment} additionally provides an illustrative customer-facing example showing how trade-up scores could be used to populate an ``Upgrade your purchase'' experience.

%% file: Sections/06_analysis.tex
\section{Discussion, Limitations, and Future Works}
\label{sec:analysis}

The component ablation further shows incremental silver-validation gains from rationale alignment and contrastive relational supervision (Appendix~\ref{app:distillation-ablation}).

\paragraph{Role of rationale supervision.}
Rationale supervision is not uniformly beneficial. 
Within the same shallow architecture, rationale supervision provides
little improvement under collapsed binary supervision but produces a
clear gain when paired with the four-class relation taxonomy (Section~\ref{ssec:reason-result}).
This suggests that the auxiliary rationale signal is most useful when the task label space can preserve the fine-grained relational distinctions encoded by the teacher.
The detailed objective ablations are provided in Appendix~\ref{app:level1}.

\paragraph{Role of product-type adaptation.}
The Level-2 results support a category-specific adaptation effect beyond the general benefit of direct expert supervision.
With the same 928 expert-labeled support examples, pooled LoRA improves AUC from $0.924$ to $0.929$, whereas product-type-specific adaptation reaches $0.940$.
Macro within-product-type AUC likewise increases from $0.910$ to $0.925$, indicating that the gain is not explained solely by cross-category score rescaling.
At the same time, reasoning-guided and label-only PT-TTT are nearly identical at $K{=}32$ ($0.941$ vs.\ $0.940$ AUC).
Taken together, these results suggest a division of labor: rationale-guided distillation is most useful for shaping the global student, whereas the Level-2 gain comes primarily from product-type-specific supervised adaptation.

\paragraph{Limitations and future work.}
Our evidence remains limited in several ways.
First, evaluation covers 29 product types and does not establish generalization to unseen categories.
Second, PT-TTT requires expert-labeled support examples and an explicit optimization step for each product type; performance may also depend on support-set composition and optimization randomness.
Third, although the pooled-adapter and macro PT-AUC controls address two alternative explanations for the observed gain, we do not fully separate product-type adaptation from simpler category-specific calibration.
Fourth, the current support/golden split is pair-disjoint but not explicitly product- or brand-disjoint; stricter entity-disjoint evaluation is therefore an important direction for future work.
Fifth, the golden benchmark is designed for controlled model comparison and may have a substantially different class prevalence from production candidate
streams; accordingly, its precision and F1 should not be interpreted directly as production positive predictive value.
Finally, the LLM comparison reflects the fixed retrieval-augmented teacher configuration evaluated in this study rather than an exhaustive optimization over prompting, retrieval, or demonstration count.
Future work should evaluate unseen-type transfer, simpler calibration and product-type-conditioned baselines, stricter entity-disjoint evaluation, and amortized adaptation that avoids per-type gradient descent.

%% file: Sections/08_conclusion.tex
\section{Conclusion}

We introduced a two-level framework for scalable trade-up identification that transfers black-box LLM supervision into an efficient embedding-pair student and then specializes it through product-type test-time training (PT-TTT).
Rationale-guided global distillation reaches AUC $0.924$ on the held-out human benchmark, while PT-TTT improves the model to AUC $0.941$ without serving-time LLM inference.
The results indicate complementary roles for the two levels: fine-grained rationale supervision is most useful during global representation learning, whereas product-type-specific expert supervision provides most of the subsequent adaptation gain.
This separation enables LLM-derived semantic supervision to be used for large-scale catalog scoring without requiring generative inference at serving time.

%% file: Sections/09_appendix.tex
\section{Supplementary Details and Results}
\label{sec:appendix}

\subsection{Data, Teacher, and Evaluation Protocol}
\label{app:data-teacher}

\subsubsection{Teacher Retrieval and Rationale Encoding}
\label{app:teacher-details}

The teacher is a frontier LLM accessed as a black box; the interface exposes
only the generated relation label and rationale text, without logits or
hidden states.

For each candidate pair, the textual information of the base product
$\base$ and candidate product $\cand$ is concatenated and encoded using the
frozen \texttt{bge-base-en-v1.5} encoder~\cite{bge_embedding}.
Query pairs are encoded using the same procedure.
Retrieval is restricted to demonstrations from the corresponding product
type, and the five examples with smallest Euclidean distance to the query
representation are selected irrespective of their relation labels.
Retrieved demonstrations are constrained to be distinct from the queried
pair.

The retrieved examples, their expert annotations, the base and candidate
product descriptions, and product-type-specific trade-up criteria are
provided to the teacher as few-shot context.
The teacher emits one four-class relation label and a concise free-text
rationale, limited to at most two sentences.

Teacher rationales are encoded once offline using the frozen
\texttt{bge-base-en-v1.5} encoder into
$d_r=768$ dimensional vectors and stored with the silver training data.
The same human-annotated demonstration pool is subsequently used as the
support-sampling pool for PT-TTT.

\subsubsection{Dataset Splits and Evaluation Statistics}
\label{app:evaluation-details}

The silver corpus contains $1{,}019{,}241$ teacher-annotated product pairs,
partitioned into training and validation sets using a stratified 90/10 split
over product types and relation classes.

The human \emph{golden} benchmark contains $8{,}352$ pairs from 29 product
types and is pair-disjoint from both the silver corpus and the teacher's
few-shot demonstration/support pool.
Silver-validation metrics are used for model, checkpoint, and hyperparameter
selection.
The resulting fixed checkpoint is evaluated on the held-out golden benchmark,
which is used only for final evaluation and analysis.

Confidence intervals are estimated using a nonparametric bootstrap with
$2{,}000$ resamples of the golden benchmark.
Pairwise model comparisons use a paired bootstrap with identical resample
indices across models, reporting the mean difference, its 95\% confidence
interval, and $P(\Delta>0)$.

AUC and AP are threshold-independent.
F1, precision, and recall are reported using the best-F1 operating threshold
selected on the silver validation set for each model, unless otherwise noted.

\subsection{Student Architecture and Training Details}
\label{app:student-details}

\subsubsection{Architecture and Hyperparameters}
\label{app:student-arch}
The student operates on one precomputed $d=768$ dimensional embedding per
product.
The base and candidate embeddings are processed by separate branches:

\begin{align}
h_b^{(0)} &= e_b, &
h_b^{(\ell)}
&=
\operatorname{SelfAttn}_b^{(\ell)}
\left(h_b^{(\ell-1)}\right),
\qquad \ell=1,\ldots,L_s, \\
h_c^{(0)} &= e_c, &
h_c^{(\ell)}
&=
\operatorname{SelfAttn}_c^{(\ell)}
\left(h_c^{(\ell-1)}\right),
\qquad \ell=1,\ldots,L_s .
\end{align}

The two branches use separate parameters, allowing side-specific
transformations for the base and candidate products.
Because each product is represented by a single embedding, these
self-attention modules operate on length-one sequences.
The attention softmax therefore contains a single element and is identically
one; in this implementation, the blocks act as parameterized transformations
rather than content-dependent token-selection mechanisms.

In parallel, an ordered base-to-candidate branch is defined as

\begin{align}
z^{(0)} &= e_b, \\
z^{(m)}
&=
\operatorname{CrossAttn}^{(m)}
\left(z^{(m-1)},e_c\right),
\qquad m=1,\ldots,L_x .
\end{align}

Because both query and key/value sides contain one embedding, the
cross-attention softmax is likewise identically one and does not perform
content-dependent selection.
The branch instead contributes an additional parameterized transformation
within the ordered base-to-candidate architecture.
Directionality of the overall classifier arises from the asymmetric
base/candidate roles, separate branch parameters, and ordered concatenation.

The branch outputs are concatenated as

\begin{equation}
h_{\mathrm{pair}}
=
\left[
h_b^{(L_s)};
h_c^{(L_s)};
z^{(L_x)}
\right]
\in\mathbb{R}^{3d}.
\end{equation}

The prediction MLP has hidden dimensions $[512,256,128]$ with ReLU
activations and dropout $0.3$, followed by a final linear layer producing
one logit in binary mode or four logits in multiclass mode.
Self- and cross-attention modules use four heads with dropout $0.1$.

For rationale-guided distillation, a linear projection maps the pair
representation into the $d_r=768$ rationale-embedding space:

\begin{equation}
\hat r_i
=
W_{\mathrm{align}}h_{\mathrm{pair}}^{(i)}
+
b_{\mathrm{align}}.
\end{equation}

This projection is used only during training and is removed at inference.
Trainable weights are initialized with Xavier-uniform initialization.

We evaluate two model capacities.
The shallow configuration uses one self-attention layer in each
side-specific branch and one cross-attention layer
($14.3$M parameters label-only; $15.5$M with the reasoning projection).
The deep configuration uses five layers of each
($64.7$M label-only; $65.9$M with the reasoning projection).

\subsubsection{Loss Definitions and Optimization Details}
\label{app:loss-details}

\paragraph{Binary task loss.}
Let $p_i=\sigma(g_i)$. The weighted focal binary objective is

\begin{equation}
\mathcal{L}_{\mathrm{bin}}
=
-\frac{1}{N}
\sum_{i=1}^{N}
\left[
w_{\mathrm{pos}}y_i^{(2)}(1-p_i)^\gamma\log p_i
+
(1-y_i^{(2)})p_i^\gamma\log(1-p_i)
\right],
\end{equation}

with $\gamma=2.0$ and validation-selected positive-class weight
$w_{\mathrm{pos}}$.

\paragraph{Four-class task loss.}
The weighted cross-entropy objective is

\begin{equation}
\mathcal{L}_{\mathrm{multi}}
=
-\frac{1}{N}
\sum_{i=1}^{N}
w_{y_i}
\log
\frac{\exp(g_{i,y_i})}
{\sum_{k=1}^{4}\exp(g_{i,k})}.
\end{equation}

\paragraph{Contrastive distillation.}
Let $\tilde r_i$ and $\tilde r_i^t$ denote the $\ell_2$-normalized
student projection and teacher rationale embedding, respectively.
We use

\begin{equation}
\mathcal{L}_{\mathrm{infonce}}
=
-\frac{1}{N}
\sum_{i=1}^{N}
\log
\frac{\exp(S_{ii})}
{\sum_{j=1}^{N}\exp(S_{ij})},
\qquad
S_{ij}
=
\frac{\tilde r_i^\top\tilde r_j^t}{\tau}.
\end{equation}

The matched pair is treated as positive and the remaining rationale
embeddings in the batch as negatives.
A relational KL objective transfers pairwise similarity structure:

\begin{equation}
\mathcal{L}_{\mathrm{kl}}
=
T^2 D_{\mathrm{KL}}
\left(P^t\,\|\,P^s\right).
\end{equation}

The complete contrastive term is

\begin{equation}
\mathcal{L}_{\mathrm{con}}
=
\mathcal{L}_{\mathrm{infonce}}
+
\lambda_{\mathrm{kl}}\mathcal{L}_{\mathrm{kl}}.
\end{equation}

We use
$\lambda_{\mathrm{align}}=0.1$,
$\lambda_{\mathrm{con}}=0.01$,
$\tau=0.07$,
$\lambda_{\mathrm{kl}}=0.05$,
and $T=1$.

\subsubsection{Optimization Details}
\label{app:optimization}

We use AdamW with learning rate $10^{-4}$, weight decay $10^{-2}$,
and $\beta=(0.9,0.999)$, with cosine learning-rate annealing to
$10^{-5}$, automatic mixed precision, and gradient-norm clipping at $1.0$.
Training uses early stopping on the monitored validation metric with patience
10.

The binary task objective uses focal BCE with $\gamma=2.0$ and the
positive-class weight used in our experiments, while the four-class task
objective uses weighted cross-entropy.
Reasoning-distillation hyperparameters are
$\lambda_{\mathrm{align}}=0.1$,
$\lambda_{\mathrm{con}}=0.01$,
$\tau=0.07$,
$\lambda_{\mathrm{kl}}=0.05$,
and $T=1$.

\subsubsection{Training Curves}
\label{app:training-curves}

\begin{figure}[t]
\centering
\includegraphics[width=\linewidth]{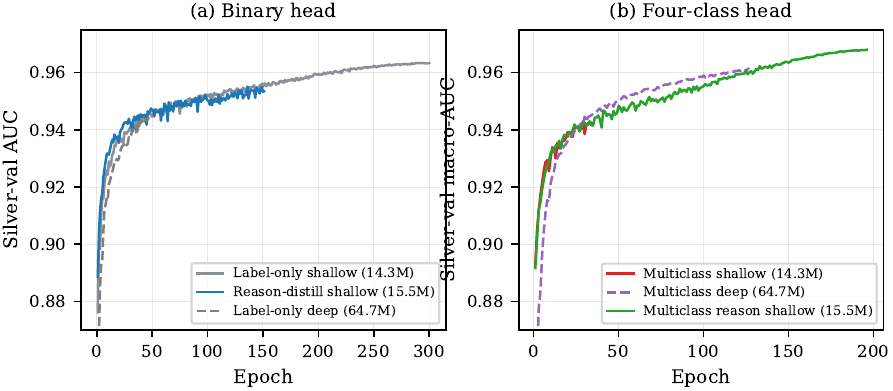}
\caption{Per-epoch silver-validation AUC.
(a) Binary head: shallow label-only and reasoning-distilled models track
closely, while the deep model performs worse.
(b) Four-class head: the deeper model performs better.}
\label{fig:training-curves}
\end{figure}

Figure~\ref{fig:training-curves} shows per-epoch silver-validation curves
for the binary and four-class heads across shallow and deep configurations.

\subsection{Additional Level-1 Distillation Results}
\label{app:level1}

\begin{figure}[!htbp]
\centering
\includegraphics[width=0.66\linewidth]{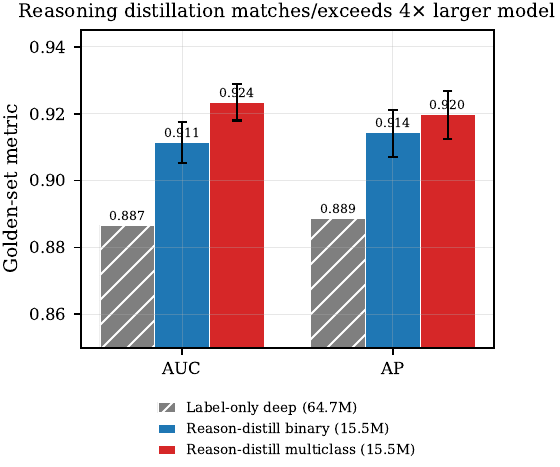}
\caption{Golden-set AUC/AP with 95\% bootstrap CIs. The shallow $15.5$M reasoning-distilled multiclass student (red) exceeds the $4\times$ larger deep label-only student (hatched grey), with non-overlapping CIs.}
\label{fig:headline}
\end{figure}

\begin{figure}[!htbp]
\centering
\includegraphics[width=0.66\linewidth]{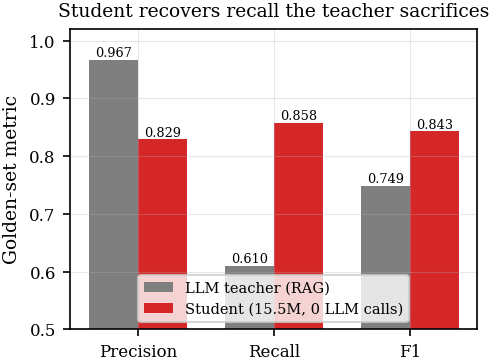}
\caption{Student vs.\ teacher on the golden set. The conservative teacher trades recall for precision; the $15.5$M student recovers recall (and a higher F1) at a modest precision cost, with zero LLM calls at inference.}
\label{fig:student-teacher}
\end{figure}

\begin{figure}[!htbp]
\centering
\includegraphics[width=\linewidth]{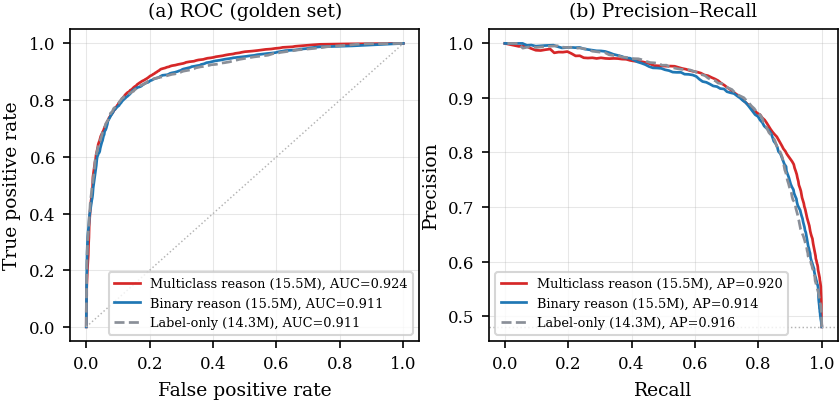}
\caption{Golden-set ROC (a) and Precision--Recall (b). The shallow multiclass reasoning-distilled student (red) dominates the binary reason and label-only baselines across operating points.}
\label{fig:roc-pr}
\end{figure}

\begin{figure}[!htbp]
\centering
\includegraphics[width=0.66\linewidth]{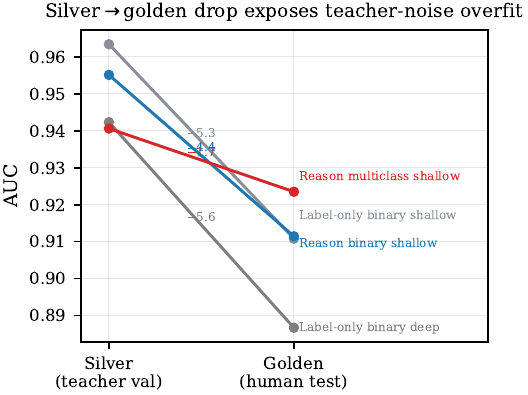}
\caption{Silver-validation vs.\ golden-test AUC. The deep binary label-only model wins on silver but drops most on golden---the signature of overfitting teacher-label noise.}
\label{fig:silver-golden}
\end{figure}

\begin{figure}[!htbp]
\centering
\includegraphics[width=0.66\linewidth]{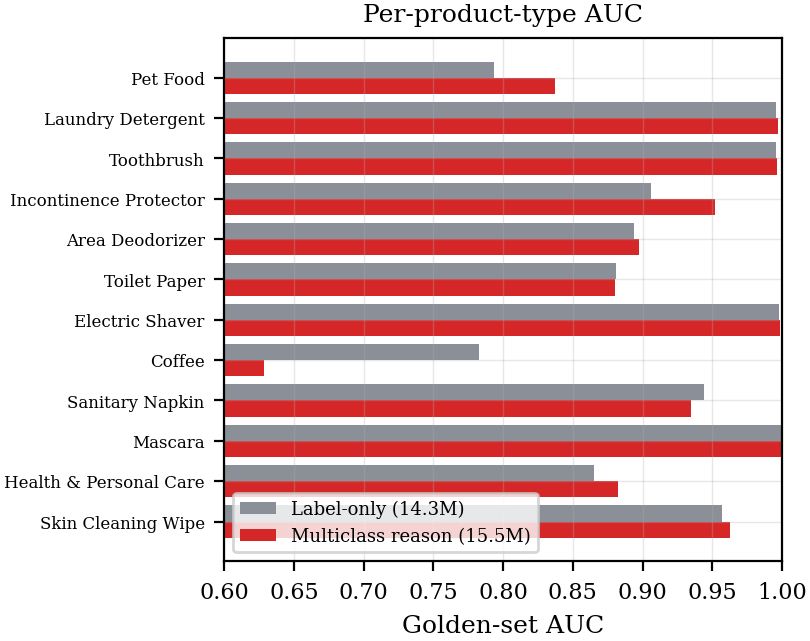}
\caption{Per-product-type golden AUC (types with ${\geq}120$ pairs in golden benchmark dataset). The multiclass reasoning-distilled student (red) matches or beats the label-only baseline in nearly all types; gains are largest where trade-up evidence is subtle.}
\label{fig:per-pt}
\end{figure}

\begin{figure}[!htbp]
\centering
\includegraphics[width=\linewidth]{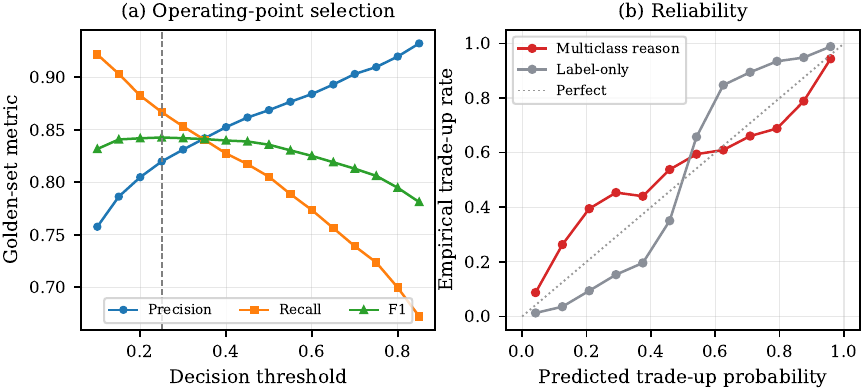}
\caption{(a) Golden-set precision/recall/F1 vs.\ threshold for the multiclass student. (b) Reliability diagram: neither student is perfectly calibrated a priori, motivating post-hoc temperature scaling before threshold selection.}
\label{fig:threshold}
\end{figure}

This appendix collects the supporting Level-1 figures deferred from Section~\ref{ssec:reason-result} for space. Figure~\ref{fig:headline} is the headline bar chart (Table~\ref{tab:reason-ablation}) with bootstrap CIs; Figure~\ref{fig:student-teacher} contrasts the student and the conservative teacher on the golden set; Figure~\ref{fig:roc-pr} shows golden ROC/PR curves; Figure~\ref{fig:silver-golden} is the silver$\rightarrow$golden generalization gap; 
Figure~\ref{fig:per-pt} is the per-product-type breakdown; and Figure~\ref{fig:threshold} shows operating-point selection and calibration.

\subsubsection{Distillation Objective Ablation}
\label{app:distillation-ablation}

\begin{figure}[!htbp]
\centering
\includegraphics[width=0.66\linewidth]{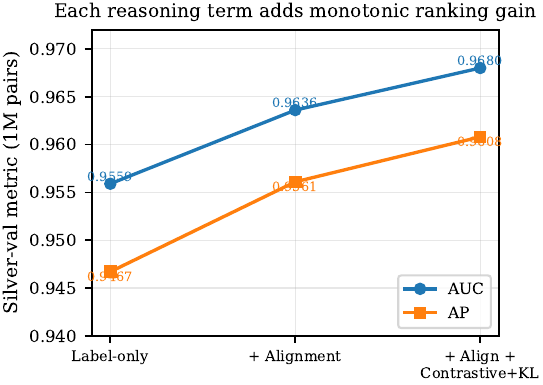}
\caption{Per-term distillation ablation on a 1M-pair dataset; metrics on the silver-validation portion): adding the alignment loss, then the contrastive (InfoNCE$+$KL) loss, each yields a monotonic ranking-quality gain (AUC $0.956{\rightarrow}0.964 {\rightarrow} 0.968$).}
\label{fig:ablation}
\end{figure}

\paragraph{The distillation objective: aligning \emph{and} contrasting rationale embeddings.}
Our Level-1 objective is a core component of the system.  Since the production student is embedding-only and never decodes text, we cannot supervise it on rationale \emph{tokens} as generative rationale-distillation methods do~\cite{hsieh2023distilling,ho2023reasoning}. We instead transfer the rationale through three complementary channels on the pair representation. 
The \emph{alignment} loss~\cite{romero2015fitnets} trains the student's projected representation to predict the teacher's rationale embedding, providing a per-example target that captures the evidence relevant to each product pair.
But pointwise alignment alone under-constrains the geometry: two pairs with different trade-up reasons can be pulled toward nearby rationale vectors. 
The \emph{contrastive} InfoNCE term~\cite{oord2018cpc,tian2020crd} repairs this by making each pair's representation discriminate its own rationale from the other rationales in the batch, preserving relational structure; the \emph{relational-KL} term~\cite{park2019relational,hinton2015distilling} further matches the teacher's full pairwise-similarity distribution rather than just the top match. The per-term ablation (Figure~\ref{fig:ablation}) confirms the channels are additive: silver AUC climbs monotonically $0.956{\rightarrow}0.964{\rightarrow}0.968$ (AP $0.947{\rightarrow}0.956{\rightarrow}0.961$) as alignment and then contrastive are added, so each earns its place. 
All three components are used only during training and removed at inference. As a result, reasoning transfer introduces no serving overhead; the deployed student retains only the improved pair representation.

\FloatBarrier
\subsection{Product-Type Test-Time Training Details}
\label{app:ttt}

\subsubsection{Adapter Parameterization and Support Protocol}
\label{app:ttt-optimization}
For each adapted linear layer $W$, LoRA parameterizes the update as
\begin{equation}
W'
=
W+\frac{\alpha}{\rho}BA,
\qquad
B\in\mathbb{R}^{d_{\mathrm{out}}\times\rho},
\quad
A\in\mathbb{R}^{\rho\times d_{\mathrm{in}}}.
\end{equation}

We use rank $\rho=8$ and scaling factor $\alpha=16$.
Adapters are inserted into the linear layers of the classification head
and the reasoning-projection layer $W_{\mathrm{align}}$, for approximately
67K trainable parameters in total.
The original global-student parameters remain frozen.

Support sets are sampled in a class-balanced manner from the
expert-annotated support pool.
Each product type is adapted independently from the same zero-delta LoRA
initialization.
The label-only and reasoning-guided variants use identical support examples
and adapter architecture, differing only in whether the rationale-alignment
term is included.

\FloatBarrier
\subsubsection{Support-Size Sensitivity}
\label{app:ttt-support}

\begin{figure}[!htbp]
\centering
\includegraphics[width=\linewidth]{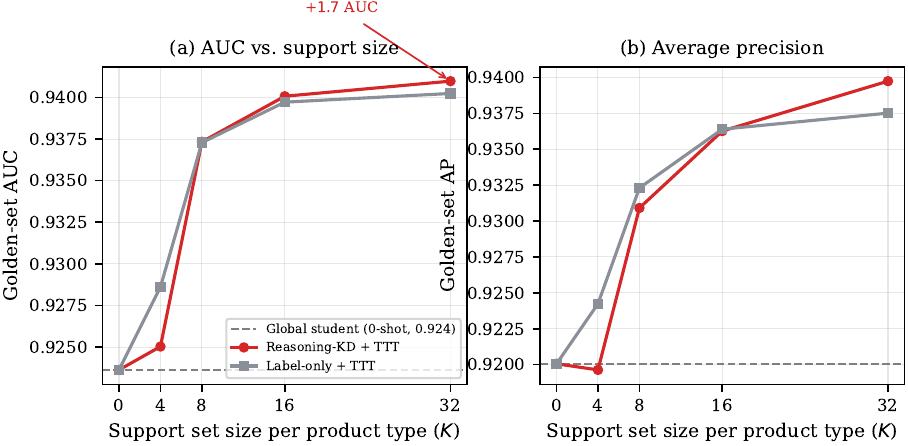}
\caption{PT-TTT vs.\ support set size $K$. 
The global student without adaptation ($K{=}0$) is the dashed baseline. 
AUC (a) and AP (b) rise monotonically to $+0.017$/$+0.020$ at $K{=}32$.}
\label{fig:ttt-curve}
\end{figure}

\FloatBarrier
\subsubsection{Adaptation-Cost Scaling}
\label{app:ttt-cost}

\begin{figure}[!htbp]
\centering
\includegraphics[width=0.62\linewidth]{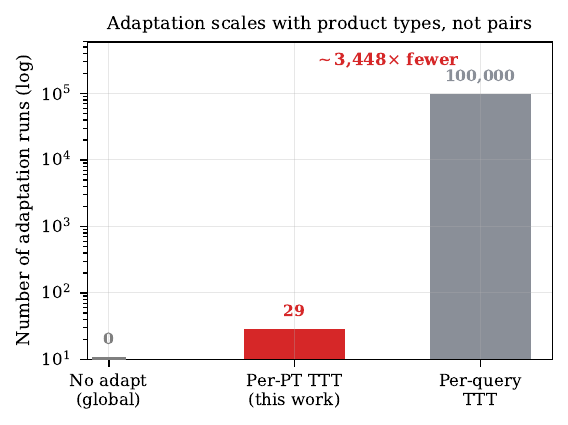}
\caption{\textbf{Adaptation-cost scaling.}
For an illustrative workload of $Q{=}100{,}000$ query pairs across
$P{=}29$ product types, PT-TTT requires $P{=}29$ adapter fits, whereas
per-query adaptation requires $Q{=}100{,}000$ fits ($3{,}448\times$ more).
Thus, PT-TTT adaptation scales with product types rather than query pairs.}
\label{fig:ttt-cost}
\end{figure}




\FloatBarrier
\subsection{Scalability}
\label{app:deployment}

\begin{figure}[!htbp]
    \centering
    \includegraphics[width=\linewidth]{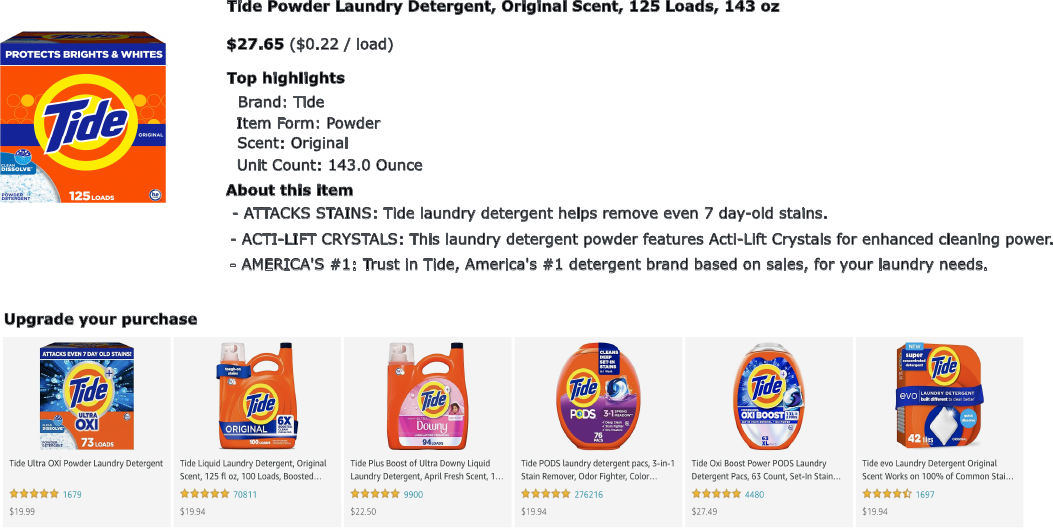}
    \caption{\textbf{Customer-facing recommendation widget.}
    Trade-Up candidate products are ranked by their predicted trade-up scores and
    presented as trade-up alternatives for the base product.}
    \label{fig:widget}
\end{figure}

To construct the customer-facing recommendation widget, we first group candidate pairs by base product and compute a trade-up score for each associated candidate offline. 
The resulting candidates are then reranked according to the desired presentation strategy. 
For example, candidates may be ordered by increasing trade-up strength to present more accessible upgrades first, or reranked using additional signals such as star rating, customer relevance, or price. 
As illustrated in Figure~\ref{fig:widget}, the customer is viewing a powder laundry detergent. 
The ``Upgrade your purchase'' widget presents ranked trade-up alternatives spanning several product formats, including enhanced-cleaning powders, liquid detergents, unit-dose pods, Power PODS, and EVO laundry tiles. 
All recommendations preserve the same underlying shopping mission, namely, laundering clothes, while offering potential improvements in cleaning performance, dosing convenience, formulation concentration, or sustainability-related product attributes. 
For example, liquid detergents may provide easier pretreatment and dispensing, whereas pods and tiles offer pre-measured dosing and reduced handling.